# From General Reasoning to Domain Expertise: Uncovering the Limits of Generalization in Large Language Models

Dana Alsagheer, Yang Lu, Abdulrahman Kamal, Omar Kamal, Mohammad Kamal,
Nada Mansour, Cosmo Yang Wu, Rambiba Karanjai, Sen Li, Weidong Shi
University of Houston
dralsagh@cougarnet.uh.edu

## Abstract

Recent advancements in Large Language Models (LLMs) have demonstrated remarkable capabilities in various domains. However, effective decision-making relies heavily on strong reasoning abilities. Reasoning is the foundation for decision-making, providing the analytical and logical framework to make sound choices. Reasoning involves analyzing information, drawing inferences, and reaching conclusions based on logic or evidence. Decision-making builds on this foundation by applying the insights from reasoning to select the best course of action among alternatives. Together, these processes create a continuous cycle of thought and action aimed at achieving goals effectively. As AI technology evolves, there is a growing trend to train LLMs to excel in general reasoning. This study explores how the general reasoning capabilities of LLMs connect to their performance in domain-specific reasoning tasks.

**ACM Reference Format:**
Dana Alsagheer, Yang Lu, Abdulrahman Kamal, Omar Kamal, Mohammad Kamal,, Nada Mansour, Cosmo Yang Wu, Rambiba Karanjai, Sen Li, Weidong Shi, University of Houston, dralsagh@cougarnet.uh.edu . 2025. From General Reasoning to Domain Expertise: Uncovering the Limits of Generalization in Large Language Models. In . ACM, New York, NY, USA, 9 pages. https://doi.org/10.1145/nnnnnnn.nnnnnnn

## 1 Introduction

Large Language Models (LLMs) have demonstrated remarkable capabilities across specialized, knowledge-intensive domains. Recent research indicates that LLMs can perform at or near expert-level proficiency in law, medicine, scientific research, and education [13, 32]. For instance, OpenAI's 2024 technical report shows that GPT-4 ranked in the 90th percentile on a simulated Multistate Bar Examination (MBE), using established preparatory materials such as BarBri [8, 53, 67]. These advances raise essential questions regarding the cognitive nature of LLMs. As AI encroaches upon domains traditionally reserved for human expertise, it evokes admiration and concern. We must ask: Are these models simply simulating human intelligence, or do they represent a fundamentally new form of cognition?

Unlike human cognition, which arises from complex biological and neurological processes, LLMs operate through algorithmic and statistical architectures. Their design enables large-scale computation and pattern recognition well beyond human capacity. In particular, some LLMs exhibit emergent behaviors - unexpected capabilities that arise on a scale - that suggest a form of reasoning or creativity not explicitly programmed [66]. These observations prompt serious consideration of whether LLMs mimic human cognition or constitute a distinct cognitive paradigm.

Through increased model parameters, massive datasets, and computational power, scaling has significantly improved language fluency and contextual understanding [36]. Yet it remains unclear whether such scaling leads to genuine advancements in reasoning or rational decision-making. Furthermore, a critical open question persists: Do improvements in one domain transfer to others, or are LLMs inherently limited to narrow task proficiency?

As LLMs are increasingly integrated into high-stakes decision-making—such as autonomous vehicles, healthcare, and legal systems—the demand for models capable of robust, generalizable reasoning grows more urgent. These systems must evolve beyond assistive tools and demonstrate the capacity to act autonomously and rationally. This work is motivated by a central research question: Do improvements in domain-specific performance (e.g., legal reasoning) translate into enhanced domain-general reasoning? We explore this question by examining the intersection of specialized task performance and broader cognitive ability in LLMs.

## 2 Contributions

This paper makes the following key contributions to the evaluation and understanding of Large Language Models (LLMs) in high-stakes, cognitively demanding domains:

- **Empirical Evaluation of Legal Expertise:** We assess the performance of state-of-the-art LLMs on a simulated Multistate Bar Examination (MBE), using authentic BarBri questions. This provides a rigorous benchmark for evaluating legal reasoning across multiple subfields.
- **Assessment of Domain-General Reasoning:** Beyond legal tasks, we examine whether LLMs demonstrate rational thinking by applying classical cognitive psychology instruments, including the Wason Selection Task and Conjunction Fallacy Test.
- **Analysis of Cognitive Transfer:** We analyze whether improvements in domain-specific performance correlate with gains in general reasoning ability. Our findings suggest limited transfer, highlighting the tendency of LLMs toward narrow, task-specific learning.

- **Human-Centered Auditing Framework:** We propose a two-step evaluation methodology incorporating expert-level benchmarking and non-expert user assessments, offering insight into actual and perceived LLM competence.
- **Implications for Alignment and Safety:** Our results reveal that high performance in specialized domains does not imply consistent or rational reasoning. These findings underscore the importance of robust auditing for the safe deployment of LLMs in real-world applications.

Together, these contributions illuminate the cognitive boundaries of current LLMs and highlight the need for human-centered, multi-faceted evaluation frameworks to guide their responsible development and use.

## 3 Related Work
### 3.1 The Great Rationality Debate in Cognitive Science

Understanding the cognitive abilities of large language models (LLMs) requires a foundation in how rationality is defined and debated within human cognition. The **Great Rationality Debate** centers on whether human reasoning aligns with normative standards such as logic, probability theory, and expected utility theory [50, 59]. These standards serve as benchmarks for evaluating human and machine decision-making.

Pioneering research by Tversky and Kahneman exposed how humans often violate these norms through heuristic-driven errors, such as the availability heuristic, representativeness, and anchoring [34, 62, 63]. These findings gave rise to the "heuristics and biases" paradigm, casting humans as "predictably irrational." Simon's concept of *bounded rationality* explains such deviations as results of limited memory, attention, and time [55, 56].

Conversely, Gigerenzer and Todd introduced the framework of *ecological rationality*, proposing that heuristics can be adaptive in real-world contexts [24, 60]. This view challenges the notion that normative models should serve as universal cognitive standards.

Dual-process theories further refine this debate. They distinguish between *System 1*, the fast and intuitive process, and *System 2*, the slow and deliberate one [21, 34]. System 2 is more closely associated with normative reasoning, but it requires cognitive effort and is not always activated unless a situation demands it.

Research also highlights the contextual and cultural variability of reasoning. Nisbett et al. and Henrich et al. demonstrated significant cross-cultural differences in cognition, calling into question the generalizability of Western psychological findings [29, 49]. Stanovich introduced a distinction between algorithmic intelligence and reflective reasoning, noting that rationality is not just about cognitive power, but also about epistemic values and metacognition [57]. Meanwhile, Cosmides and Tooby argued that reasoning is governed by domain-specific modules shaped by evolution [17, 61].

Together, these perspectives complicate simplistic notions of rationality and underscore the need for contextualized evaluation—an insight that has direct implications for auditing AI models.

### 3.2 Machine Psychology: Cognitive Approaches to Artificial Intelligence

The emerging field of *Machine Psychology* investigates the cognitive behaviors of artificial intelligence systems, especially LLMs, through the lens of psychology and experimental cognitive science [27]. This behavioral approach moves beyond architectural interpretability by treating models as experimental subjects capable of producing measurable reasoning patterns.

Inspired by human psychological methods, researchers have applied cognitive reflection tests, logical reasoning tasks, and behavioral probes to study whether LLMs exhibit human-like decision-making [11, 18, 39]. While these models often demonstrate impressive linguistic fluency, they fall short in tasks requiring abstract reasoning or the suppression of heuristics.

Role-based prompting further reveals LLMs' behavioral plasticity. When prompted as domain experts, LLMs can exhibit surface-level competence in law or medicine but fail to generalize these skills to unfamiliar domains [5, 40, 68]. Such limitations raise doubts about claims of general intelligence and highlight the challenges of domain transfer.

Emergent capabilities such as in-context learning—where models appear to adapt to prior prompts—have been debated. Some interpret these as signs of reasoning ability, while others suggest they reflect superficial pattern matching rather than deep understanding [44, 69].

The question of how to benchmark AI cognition remains contentious. While some advocate for human-inspired metrics, others argue that these benchmarks risk anthropomorphizing systems that fundamentally differ from biological cognition [9, 37, 42, 45].

### 3.3 Biases and Limitations in LLM Reasoning

Despite their growing capabilities, LLMs remain prone to significant reasoning flaws and cognitive biases, mirroring—but not necessarily replicating—human limitations. Studies have documented that LLMs exhibit confirmation bias, anchoring, and overconfidence when responding to prompts involving ambiguity or uncertainty [48, 70].

These biases pose serious risks in high-stakes environments such as law, healthcare, and public policy. As Hagendorff argues, biased outputs are not mere technical glitches but emergent behaviors of complex systems shaped by training data and task framing [26]. Similarly, Doshi-Velez and Kim, and Ouyang et al., emphasize that mitigating these biases is essential for trustworthy AI, especially when systems are involved in consequential decision-making [20, 54].

Research into the generalization limitations of LLMs suggests that while models like GPT-4 may display high performance in structured benchmarks, they often struggle with cross-domain transfer and consistency [12, 14, 38]. This undermines the assumption that performance in narrow tasks equates to general intelligence.

Ultimately, bias in LLMs is not simply a technical issue, but a cognitive one. Understanding and mitigating these limitations requires drawing on interdisciplinary insights from psychology, cognitive science, and ethics.

Table 1: Summary of Large Language Models (LLMs) by size, availability, and citations (sorted by size).

| Model | Size (Parameters) | Availability | Notes | Cite |
| --- | --- | --- | --- | --- |
| Mistral | 7B-12B | Free | Lightweight and efficient. | [3] |
| Claude 2 | ~52-100B | Free and Paid | Safety-centric, developed by Anthropic. | [6] |
| Llama 3 (70B) | 70B | Free | Openly available under Meta's licensing terms. | [2] |
| Gemini 1 | ~70B | Free and Paid | Competitive with ChatGPT-4 in performance. | [19] |
| ChatGPT-3.5 | ~175B | Free and Paid | Reliable for everyday tasks. | [51] |
| ChatGPT-4 | Estimated 1-1.76T | Paid | High performance in reasoning tasks. | [52] |

## 3.4 Rationality and its Importance in AI Systems

Rationality is making decisions, solving problems, and achieving goals through structured reasoning, critical analysis, and evidence-based actions. It is divided into two key types: *theoretical rationality*, which governs beliefs and logical reasoning, and *practical rationality*, which pertains to actions aimed at achieving desired outcomes [21]. These two types are essential for tasks requiring precision and logical structure, such as legal analysis, scientific inquiry, and mathematical problem-solving.

Rationality also plays a critical role in artificial intelligence systems, ensuring reliable and structured performance. Large Language Models (LLMs) are increasingly evaluated for their capacity to demonstrate rational reasoning in specialized tasks, such as legal analysis and logical problem-solving. Fine-tuning LLMs on domain-specific datasets has shown notable improvements in producing rational and context-aware outputs. However, these enhancements often remain confined to the specific training domain, raising questions about their ability to perform consistently across multiple contexts [23, 66].

The importance of rationality in AI becomes particularly evident as LLMs are integrated into high-stakes systems, including healthcare, finance, and autonomous technologies [20, 54]. In such domains, rational reasoning ensures the AI's outputs are accurate and effective. Rationality in AI can be further categorized into *epistemic rationality*, which focuses on producing outputs grounded in truth and accurately reflecting reality, and *instrumental rationality*, which ensures that the system's actions achieve specific goals efficiently. For instance, in a medical setting, an LLM assisting with diagnoses must exhibit epistemic rationality to interpret clinical data accurately and instrumental rationality to recommend the safest and most effective treatment plans.

This growing reliance on LLMs highlights the need for strategies to evaluate and enhance their rationality, particularly in domain-specific applications. Ensuring logical consistency and task-specific rationality in AI systems is essential for their adoption in critical decision-making processes.

## Methodology

This study examines the capabilities of large language models (LLMs) across narrow domain-specific tasks and broader reasoning abilities. We aim to evaluate whether these abilities are interconnected and how model size influences their performance. Our approach examines the connection between domain-specific skills like legal reasoning and broader cognitive abilities, including rationality and logical decision-making, without assuming a unified framework linking these domains.

We evaluated six prominent large language models (LLMs) of varying parameter sizes: OpenAI's ChatGPT-3.5 and ChatGPT-4, Google's Gemini 1.0, Meta's Llama 3, Anthropic's Claude 2, and Mistral AI. The models ranged from Mistral's 7 billion parameters to ChatGPT-4's 175 billion parameters, allowing us to analyze the impact of scale on task performance. Details of the models and their sizes are summarized in Table 1.

### Data Collection

To assess **narrow domain-specific abilities**, we compiled a dataset of 1,000 legal questions derived from the Multistate Bar Examination (MBE). These questions span seven core legal domains: Civil Procedure, Constitutional Law, Contracts, Criminal Law, Evidence, Real Property, and Torts [8]. The questions were selected for their analytical rigor and relevance to real-world legal practice, challenging the models' capacity to interpret, analyze, and apply complex legal principles. This dataset's comprehensive scope ensures minimal bias and captures meaningful differences in models' ability to perform domain-specific tasks. Table 2 provides further definitions of the legal domains.

We utilized a diverse set of established reasoning tasks to evaluate **broader reasoning abilities**. These included the Wason Selection Task [65], the Conjunction Fallacy Task [64], the Stereotype Base Rate Neglect Task [47] and the Diagnostic Base Rate Neglect Task [15, 35], These tasks assessed models' logical reasoning, probabilistic thinking, and decision-making abilities, with 240 items providing broad coverage of rationality-based tasks. Figures 1 illustrate these tasks' key examples and comparative performance metrics. Table 3 offers detailed task definitions.

Each model was accessed through its official API using default configurations to ensure consistency. Over 30 days, we conducted five rounds of evaluation per task to account for temporal variations, using standardized prompts across trials. Results were averaged for reliability. Python-based tools with automated scoring minimized potential human bias, ensuring objective evaluation. All datasets, prompts, and scripts for data collection will be provided as supplementary material.

### Correlation Analysis of Abilities

We investigated the relationship between narrow and broader reasoning abilities by calculating **Pearson's correlation coefficient** [10]. The correlation coefficient measures the strength and direction of a linear relationship, ranging from -1 (perfect negative relationship) to +1 (perfect positive relationship), with 0 indicating no linear correlation. This method allowed us to determine whether performance improvements in one domain were associated with enhancements in the other.

Table 2: Overview of Multistate Bar Examination (MBE) Tasks and Their Definitions

| Task | Definition | Cite |
| --- | --- | --- |
| Civil Procedure | The body of law governing the processes and rules courts follow in civil lawsuits, including how cases are filed, tried, and appealed. | [4] |
| Constitutional Law | The area of law that interprets and applies the U.S. Constitution, governing the relationships between the government and individuals and the division of powers among government branches. | [1, 4] |
| Contracts | The branch of law that deals with agreements between parties, including the creation, enforcement, and breach of legally binding agreements. | [4, 22] |
| Criminal Law and Procedure | The area of law that defines criminal offenses and the legal process for prosecuting and defending against charges, including arrest, trial, and sentencing. | [4, 28] |
| Evidence | The rules and principles that govern what information can be presented in court to prove or disprove facts in a legal proceeding. | [4, 33] |
| Real Property | Land and anything permanently attached to it, such as buildings and the rights associated with land ownership. | [4, 30] |
| Torts | The area of law deals with civil wrongs or injuries, providing remedies for individuals harmed by the actions or omissions of others. | [4] |

Table 3: Definitions of Rationality Tasks

| Task | Definition | Cite |
| --- | --- | --- |
| Wason Selection Task | A task assessing deductive reasoning, wherein participants are required to select cards to evaluate the validity of a conditional rule (e.g., "If P, then Q"). This paradigm specifically measures the capacity to identify evidence that confirms or falsifies logical propositions under constrained conditions. | [15, 65] |
| Conjunction Fallacy Task | A task designed to investigate systematic violations of normative probabilistic reasoning, wherein participants erroneously ascribe a higher likelihood to the conjunction of two events than a single, broader event. This paradigm highlights the pervasive cognitive bias rooted in heuristic processing. | [64] |
| Stereotype Base Rate Neglect Task | A task that assesses the inclination to prioritize symbolic or anecdotal information over statistical base rates. It examines how reliance on heuristics, frequently influenced by stereotypical patterns, can lead to deviations from rational probabilistic reasoning. | [47] |
| Diagnostic Base Rate Neglect Task | A task is evaluating the interplay between statistical base rates and diagnostic cues, specifically probing how individuals overemphasize vivid or diagnostic details at the expense of statistically grounded reasoning. This paradigm reveals the limitations in integrating probabilistic data within decision-making frameworks. | [15, 35] |

Correlation coefficients and corresponding *p*-values were computed using Python's `scipy.stats` library, specifically the `pearsonr` function. A significance threshold of 0.05 was applied to the *p*-values to identify statistically significant relationships. This computational approach ensured the accuracy and reproducibility of our findings. The analysis code and detailed results are included as supplementary material.

## 4 Results

The study aims to go beyond simply evaluating the accuracy of Large Language Models (LLMs). It examines the relationship between legal reasoning, a specialized, domain-specific capability, and rationality, a broader cognitive ability. It explores whether advancements in specialized domains enhance general cognitive capabilities akin to human intelligence or if such progress remains limited to isolated areas. This section presents the study's findings, emphasizing three critical areas:

- The relationship between narrow and broad abilities
- The influence of model size on the interrelationship among these abilities
- Unique cognitive bias patterns that distinguish LLM intelligence from human cognition

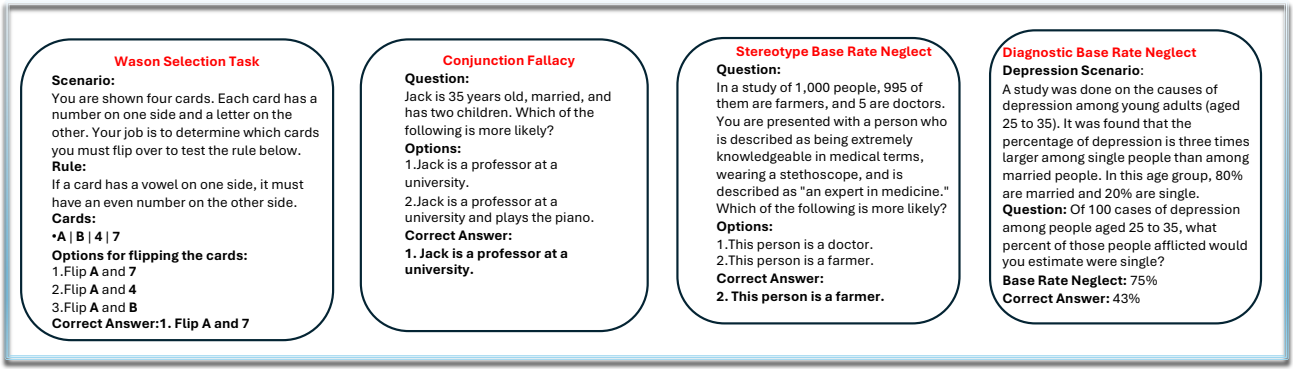

**Figure 1: Examples of diagnostic assessments designed to evaluate rational decision-making abilities include the Wason Selection and Conjunction Fallacy Task.**

These findings indicate that narrow and broad cognitive abilities are disconnected in the intelligence structure of LLMs.

## 4.1 Investigating the Correlation Between Narrow and Broad Abilities in LLMs

We inspected the potential relationship between narrow and broad abilities in Large Language Models (LLMs) by comparing performance in legal reasoning tasks (narrow abilities) and rationality tasks (broad abilities). Pearson's correlation coefficients were calculated for each model to assess the strength and significance of these relationships. The results are summarized in Table 4.

The correlation analysis revealed no statistically significant relationships between performance in legal reasoning and rationality tasks for any of the LLMs evaluated. All $p$-values exceeded the 0.05 threshold, indicating a lack of significant correlations. The Pearson correlation coefficients ranged from -0.11 to 0.29, suggesting weak or negligible relationships between the two abilities. These findings indicate that the models do not exhibit a meaningful connection between narrow and broad abilities.

In all models, the legal reasoning and rationality performance appeared to operate independently, with no evidence of generalized or transferable cognitive capacity. The lack of correlation shows the specialized nature of LLMs' abilities, with each model demonstrating strengths in specific domains without cross-domain generalization. This pattern highlights LLMs' distinct and isolated performance characteristics in these tasks.

## 4.2 The Influence of LLM Model Size on Domain-Specific and Broad Abilities

We examined whether the size of Large Language Models (LLMs), defined by the number of parameters, influences the relationship between domain-specific abilities (e.g., legal reasoning) and broader cognitive abilities (e.g., rationality). Our findings indicate that increasing model size improves performance in specific tasks but does not enhance the correlation between narrow and broad abilities.

Larger models consistently demonstrate higher scores in both narrow and broad tasks. For instance, ChatGPT-4 achieved an average score of 64.57 in legal reasoning and 61.29 in rationality. Despite these improvements, the correlation between these abilities was weak and statistically insignificant ($r = 0.14, p = 0.76$). Similarly, ChatGPT-3.5 ($r = 0.29, p = 0.53$) and Claude ($r = 0.01, p = 0.98$) achieved strong performance in individual domains but failed to show meaningful relationships between narrow and broad abilities.

Smaller models further reinforce this trend. Llama, with scores of 55.58 in legal reasoning and 38.43 in rationality, exhibited a weak negative correlation ($r = -0.11, p = 0.82$). In contrast, Mistral, with only 7 billion parameters, achieved scores of 36.51 and 31.84 in legal reasoning and rationality, respectively, and demonstrated a weak positive correlation ($r = 0.22, p = 0.64$). These results highlight the fragmented nature of LLM performance, where narrow abilities improve with model size but remain disconnected from broader abilities. Figure 2 demonstrates these results.

The findings align with prior observations in LLM research, which demonstrate that model scaling enhances task-specific performance without fostering general domain integration [25, 31, 36]. The isolated performance across domains underscores the need for innovative architectural and training strategies to address this limitation. Without such advancements, LLMs will continue to exhibit impressive yet fragmented abilities confined to specific domains.

## 4.3 Cognitive Bias Analysis: Insights into LLM Limitations and Domain-Specific Reasoning

Cognitive biases—systematic patterns of deviation from normative reasoning—are foundational to understanding both human and artificial intelligence. In humans, these biases reflect evolutionary heuristics that operate efficiently under bounded cognitive resources [48, 62]. However, when such patterns appear in Large Language Models (LLMs), they reveal critical limitations in how these systems simulate reasoning and adapt across domains [7].

Table 4: Correlation Between Legal Reasoning and Rationality Across LLMs Ranked by Model Size (Largest to Smallest)

| Model | Average Law Score | Average Rationality Score | Pearson Correlation | P-Value | Relationship |
| --- | --- | --- | --- | --- | --- |
| ChatGPT-4 | 64.57 | 61.29 | 0.14 | 0.76 | No significant relationship |
| Gemini | 41.71 | 34.43 | 0.20 | 0.67 | No significant relationship |
| ChatGPT-3.5 | 36.86 | 38.71 | 0.29 | 0.53 | No significant relationship |
| Claude | 59.57 | 67.29 | 0.01 | 0.98 | No significant relationship |
| Llama 3.2 | 55.58 | 38.43 | -0.11 | 0.82 | No significant relationship |
| Mistral | 36.51 | 31.84 | 0.22 | 0.64 | No significant relationship |

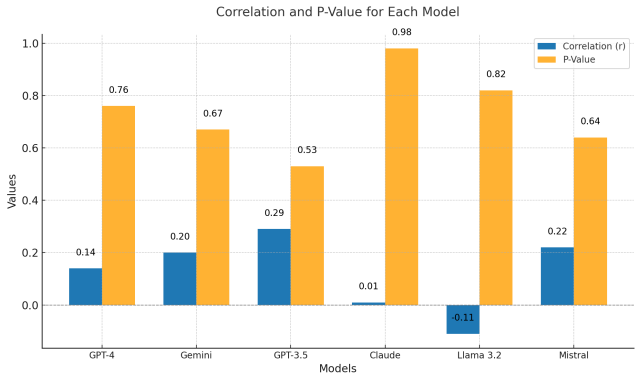

Figure 2: Larger models, like ChatGPT-4, achieve higher average scores in both narrow (legal reasoning) and broad (rationality) tasks. However, the lack of significant correlation suggests that improvements in one domain do not translate into improvements in the other. Smaller models, such as Mistral, also fail to demonstrate meaningful correlations.

Our comparative analysis of LLM performance across classic cognitive psychology tasks revealed both systematic biases and notable instabilities, highlighting the fragile nature of domain-specific reasoning in these models. LLMs, despite their proficiency in specialized contexts, frequently relied on surface-level heuristics and statistical associations rather than robust, abstract reasoning strategies.

For instance, in the **Wason Selection Task**, models exhibited a strong bias toward pattern matching over rule-based deduction. Rather than identifying the logically correct cards to test a conditional statement, LLMs defaulted to familiar linguistic structures from their training corpus, missing the deeper logical structure of the task [48]. This reflects a limitation in their ability to generalize logical reasoning outside high-frequency training distributions.

In the **Conjunction Fallacy Task**, models often selected the more intuitively compelling but statistically incorrect answer, favoring narrative coherence over probability theory. This mirrors the representativeness heuristic in human cognition, where plausibility overrides formal logic [62]. Such responses suggest that LLMs optimize for textual plausibility rather than underlying rational principles.

Similarly, in the **Base Rate Neglect Task**, LLMs demonstrated anchoring effects—prioritizing salient information (e.g., personality traits) over base rate probabilities. This behavior occurred even when the base rate was explicitly stated in the prompt, indicating a tendency to overfit on linguistically salient cues at the expense of statistical reasoning [7].

A surprising pattern emerged when comparing domain complexity. LLMs performed well on specialized tasks involving formal mathematics or legal analysis, likely due to high exposure to such structured data during training. However, they often underperformed on simpler problems requiring basic logic or probabilistic consistency, revealing an uneven cognitive profile. This discrepancy suggests that their apparent expertise in certain areas does not translate into a cohesive or generalizable reasoning system [58].

Most critically, our study uncovered a recurring issue across all evaluated models: **inconsistency in responses to identical prompts**. When the same question was asked multiple times under identical conditions, models such as ChatGPT-4 frequently returned different answers. This test-retest inconsistency occurred even in tasks with unambiguous correct answers and no temperature or sampling variation. These fluctuations underscore a fundamental instability in LLM reasoning—despite their deterministic architectures, their internal state and sampling dynamics allow for reasoning pathways that lack coherence across iterations.

Taken together, our cognitive bias analysis reveals that LLMs, while capable of high performance in isolated domains, are limited by their reliance on context-specific heuristics, inability to generalize across domains, and lack of internal consistency. These findings challenge the assumption that task-specific success implies broader cognitive competence, and underscore the need for new evaluation frameworks that incorporate both *bias sensitivity* and *reasoning stability* as core indicators of general intelligence [7].

## 5 Discussion

Intelligence can be defined as the ability of an entity to select the optimal course of action in any given situation to achieve its primary objective while also creating and pursuing sub-goals as needed to reach that objective. Extending this definition to AI models implies an expectation that they can identify the optimal actions to accomplish a goal across various circumstances and autonomously generate sub-goals to facilitate the process.

We anticipate that AI models will achieve a set of objectives given by humans, which is why we adopt a goal-oriented approach. However, actions that initially appear most advantageous for achieving the ultimate goal may prove less effective over the long term. Therefore, AI models should prioritize accomplishing their goals through strategic decision-making rather than focusing solely on immediate benefits.

To attain human-level AI, models should mirror human mental processes. They are expected to replicate core human cognitive functions such as perception, memory, attention, decision-making, social intelligence, etc. AI models are anticipated to comprehend and justify both their own actions and decisions and those of other agents.

However, this anticipation remains theoretical. Many unanswered concerns and ongoing discussions regarding its viability and moral ramifications exist. How should we determine whether large language models (LLMs) can achieve human-level intelligence? There are multiple ways to assess AI models. In this research, we choose the indicators of the linkage between domain-specific reasoning and general reasoning. Applying intelligence tests developed for humans to LLMs raises questions about how to interpret the results. What can we infer from the high scores LLMs achieve? We must be very cautious in concluding and acknowledging the limitations and differences between human and artificial intelligence.

There is an increasing emphasis on optimizing LLMs performance in specific tasks or domains, often driven by the mistaken belief that such targeted improvements will lead to enhancements in general cognitive abilities. Our analysis of over 2,024 publications identified approximately 1,025 research efforts focused on enhancing specific domain capabilities. A detailed list of these studies is provided in the supplementary materials, emphasizing their predominant focus on particular cognitive abilities. As shown in Figure 3, most publications focus on advancing broad abilities, such as Fluid Intelligence (Gf) and Crystallized Intelligence (Gc) [43], or enhancing narrow specific domain skills. However, our findings reveal a critical gap: no significant connection was observed between narrow and broad abilities. This observation suggests that isolated improvements in domain-specific reasoning do not contribute to the development of genral reasoning. It prompts a re-evaluation of the value of such efforts without a holistic understanding of cognitive interdependence.

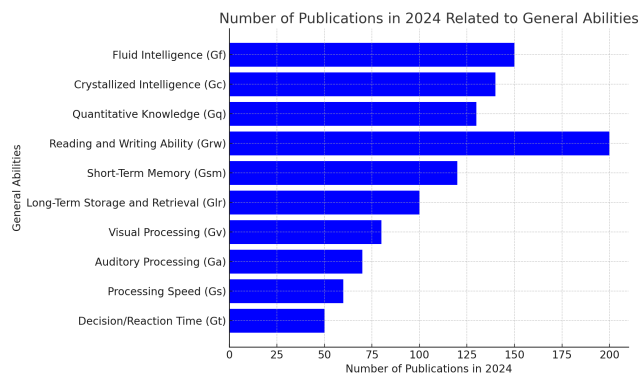

**Figure 3: Number of publications in 2024 focusing on enhancing broad cognitive abilities in Large Language Models (LLMs).**

Our research results show that LLMs lack this integrative growth. While they may achieve impressive results in standardized tests like the MBE [9, 46], they fail to reflect on the general reasoning. This misalignment diverts valuable research efforts toward optimizing LLM performance on standardized exams rather than exploring meaningful metrics for assessing general intelligence [16, 41].

Additionally, LLMs often demonstrate a striking inconsistency in their abilities, excelling in complex tasks such as advanced mathematical reasoning while struggling with simpler tasks like basic arithmetic and numerical operations. This discrepancy highlights the lack of a cohesive connection between abilities within LLMs, as success in one area does not translate into improved performance in other domain tasks.

Analyzing the errors made by LLMs further highlights this issue. Human errors are frequently attributable to biases or misinterpretations, or emotional factors. In contrast, the errors in large language models (LLMs) typically originate from limitations in training data, contextual misunderstandings, or the application of rigid algorithmic reasoning. This divergence in error patterns necessitates the creation of evaluation frameworks specifically tailored to LLMs. These frameworks should acknowledge the unique capabilities and limitations of these models, avoiding the misleading comparison of LLM performance to human cognition. Such tailored evaluations are crucial for accurately assessing LLM strengths and weaknesses, guiding future development, and ensuring responsible deployment.

While LLMs excel at specific tasks, replicating human general intelligence remains a significant hurdle. Their limited ability to generalize poses major challenges for high-stakes fields like law, medicine, and scientific research, which demand adaptability, contextual understanding, and logical consistency. Overcoming these limitations is crucial for developing next-generation AI systems that can move beyond narrow applications and achieve human-like adaptability in diverse real-world scenarios.

## 6 Conclusion

This study critically examined the cognitive capabilities of Large Language Models (LLMs), focusing on whether advancements in narrow, domain-specific tasks—such as legal reasoning—translate into broader, domain-general reasoning abilities, including rationality and consistency. While LLMs like GPT-4 exhibit impressive performance on high-stakes benchmarks such as the Multistate Bar Examination (MBE), our analysis reveals that such success is often isolated and does not imply the presence of a generalizable reasoning system.

Across all evaluated models and tasks, we found no statistically significant correlation between performance in specialized legal domains and performance on classical cognitive tasks assessing logical reasoning and probabilistic judgment. This lack of cognitive transferability challenges assumptions that scaling or fine-tuning alone can produce general intelligence. Even the largest models demonstrated fragmented capabilities: excelling in tasks aligned with training data distributions while struggling in simpler, foundational reasoning problems.

Moreover, our cognitive bias analysis uncovered systemic limitations in LLMs' reasoning, including reliance on representativeness heuristics, anchoring effects, and failures in logical deduction. Most concerning, however, was the pervasive inconsistency we observed—models frequently provided divergent answers to identical questions under controlled, deterministic conditions. This test-retest instability undermines the reliability of LLM outputs in settings where consistency is paramount.

Taken together, our findings suggest that current LLMs operate as collections of narrow capabilities rather than cohesive cognitive agents. Their performance is best described as pattern-based mimicry rather than true understanding, lacking the integrative structure observed in human intelligence. As such, high task accuracy should not be conflated with cognitive competence or rational decision-making.

To advance toward generalizable, trustworthy AI, we argue for a paradigm shift in how model intelligence is evaluated. Future research should prioritize not only domain-specific benchmarks but also frameworks that assess reasoning stability, cognitive transfer, and cross-domain coherence. Human-centered auditing, cognitive diagnostics, and interdisciplinary evaluation protocols are essential for understanding and guiding the next generation of LLMs beyond narrow specialization and toward broader, more adaptable intelligence.